%% file: Main.tex
\documentclass[runningheads]{llncs}
\usepackage{graphicx}

\usepackage{tikz}
\usepackage{comment}
\usepackage{amsmath,amssymb} 
\usepackage{color}

\usepackage[width=122mm,left=12mm,paperwidth=146mm,height=193mm,top=12mm,paperheight=217mm]{geometry}
\usepackage{xcolor}
\usepackage{booktabs}
\usepackage{multirow}
\usepackage{subcaption}
\usepackage{wrapfig}

\usepackage{pifont}

\newcommand{\bbR}{{\mathbb{R}}}

\begin{document}
\pagestyle{headings}
\mainmatter
\def\ECCVSubNumber{1496}  

\title{Transforming Multi-Concept Attention into Video Summarization} 

\titlerunning{MC-VSA}
%
\author{Yen-Ting Liu$^{1}$\and
Yu-Jhe Li$^{2 }$ \and
Yu-Chiang Frank Wang$^{1}$}
\authorrunning{Liu \emph{et~al.}}
%
\institute{$^{1}$National Taiwan University, Taipei, Taiwan \\
$^{2}$Carnegie Mellon University, Pittburgh, PA, USA\\
\email{\url{r06942114@ntu.edu.tw}}\hspace{4.0mm}\email{{yujheli}@cs.cmu.edu}\hspace{4.0mm}\email{{ycwang}@ntu.edu.tw}
}
\maketitle

\input{0_abstract.tex}

\input{1_introduction.tex}

\input{2_related_work.tex}

\input{3_method.tex}

\input{4_experiment.tex}

\input{5_conclusion.tex}

\clearpage
%
%
\bibliographystyle{splncs04}
\bibliography{egbib}
\end{document}

%% file: 0_abstract.tex
\begin{abstract}
    Video summarization is among challenging tasks in computer vision, which aims at identifying highlight frames or shots over a lengthy video input. In this paper, we propose an novel attention-based framework for video summarization with complex video data. Unlike previous works which only apply attention mechanism on the correspondence between frames, our \textit{multi-concept video self-attention (MC-VSA)} model is presented to identify informative regions across temporal and concept video features, which jointly exploit context diversity over time and space for summarization purposes. Together with consistency between video and summary enforced in our framework, our model can be applied to both labeled and unlabeled data, making our method preferable to real-world applications. Extensive and complete experiments on two benchmarks demonstrate the effectiveness of our model both quantitatively and qualitatively, and confirms its superiority over the state-of-the-arts.
\keywords{video summarization, attention model, multi-concept}
\end{abstract}

%% file: 1_introduction.tex
\section{Introduction}
\label{sec:intro}

\input{Figures/1_teaser.tex}

Video summarization \cite{chao2015large,gygli2014creating,gygli2015video,zhang2016summary} aims at identifying highlighted video frames or shots, which is among the challenging tasks in computer vision and machine learning. Real-world applications such as video surveillance, video understanding and retrieval would benefit from successful video summarization outputs.
To address this challenging task, several deep learning-based models~\cite{zhang2016video,jung2018discriminative,mahasseni2017unsupervised,zhang2018retrospective,zhou2018deep} employing long short-term memory (LSTM) cells~\cite{LSTM1997} have been recently proposed. However, the use of such recurrent neural network (RNN) based techniques might fail if the length of the input video is long~\cite{yue2015beyond}. Therefore, even the training video data are with ground-truth labels, there is no guarantee that RNN-based models would achieve satisfactory results using the last output state.
To address the aforementioned issue, several approaches (also based on deep learning) are proposed~\cite{zhao2017hierarchical,zhao2018hsa,rochan2018video}. For example,~\cite{zhao2017hierarchical,zhao2018hsa} advances hierarchical structure LSTMs to capture longer video, which is shown to be able to handle video with longer lengths. \cite{rochan2018video} proposes SUM-FCN which considers CNN-based semantic segmentation model to deal with videos while alleviating the above concern. Yet, these existing techniques might not exhibit capabilities in modeling the relationship between video frames, since they generally treat each frame equally important. Thus, their summarization performance might be limited.

To advance the attention mechanism for video summarization, a number of methods are recently proposed~\cite{fu2019attentive,ji2017video,ji2017attention}.  With the ability of learning importance weights across all video frames, attention-based models are expected to be more robust while it is still able to tackle lengthy video as well as the video semantic meanings. For example, ~\cite{ji2017video} firstly proposes attentive video summarization model (AVS) for improved attention on summarization tasks. ~\cite{ji2017attention} also employs attention models for properly identifying video shot boundaries. Nevertheless, these attention-based methods might not generalize to general videos with complex content information, since they typically perform attention on pre-selected feature representations or latent spaces. To make a summarization model more robust to real-world video, one needs to better observe and relate the temporal-concept information within and across video frames, rather than exclusively attend correlation between frames in the video.

In this paper, we propose a novel attention-based deep learning framework for video summarization. With the goal to attend temporally and concept informative feature for summarization in the sense, we present a \textit{multi-concept video self-attention (MC-VSA)} model in a discriminative learning mechanism. Based on the idea of \cite{vaswani2017attention}, we add the multi-head attention mechanism in our model to transform input video frames features into different subspaces. Different from previous attention model~\cite{fu2019attentive,ji2017video,ji2017attention}, this allows us to exploit a variety of visual appearances during the attention process, and thus identify visual concept informative regions across frames for both \textit{summarization} and \textit{video semantic consistency} purposes, which we call multi-concept attention cross whole video time step (temporal and concept attention).

Take an example as illustrated in Fig.~\ref{fig:picture1}, it would be desirable to be able to extract different visual concepts corresponding to different semantics or objects with the highlight guidance, so that the joint attention across these concepts would allow satisfactory video summarization outputs.  More importantly, our learning framework can be generalized well in a semi-supervised setting, i.e., only a number of training data are with ground-truth labels.
More details of our proposed framework will be presented in Sec.~\ref{sec:Method}. In addition, we found that current evaluation protocol using pre-defined procedure has some problems~\cite{song2015tvsum,gygli2015video} (e.p., random summaries outperform even the human generated summaries in leave-one-out experiments), which are mentioned in~\cite{re2019}. Therefore, our quantitative experiment and ablation study are based on both the current~\cite{song2015tvsum,gygli2015video} and the new~\cite{re2019} evaluation protocol.

The contributions of this paper is highlighted as follows:
\begin{itemize}
    \item We present a multi-concept video self-attention model (MC-VSA) aims at attending temporally and concept informative features via transforming input video frames in different subspaces, which is beneficial to video summarization purposes.
    \item We are among the first to propose the attention based framework that observes semantic consistency between input videos and learned summarization outputs, which allows the video summarization model can be generalized in semi-supervised settings.
    \item Experimental results on benchmark datasets confirm that our proposed method achieves favorable performances against the state-of-the-art approaches in two evaluation protocols.
\end{itemize}

%% file: Figures/1_teaser.tex
\begin{figure}[t]
\begin{center}
\includegraphics[width=0.7\linewidth]{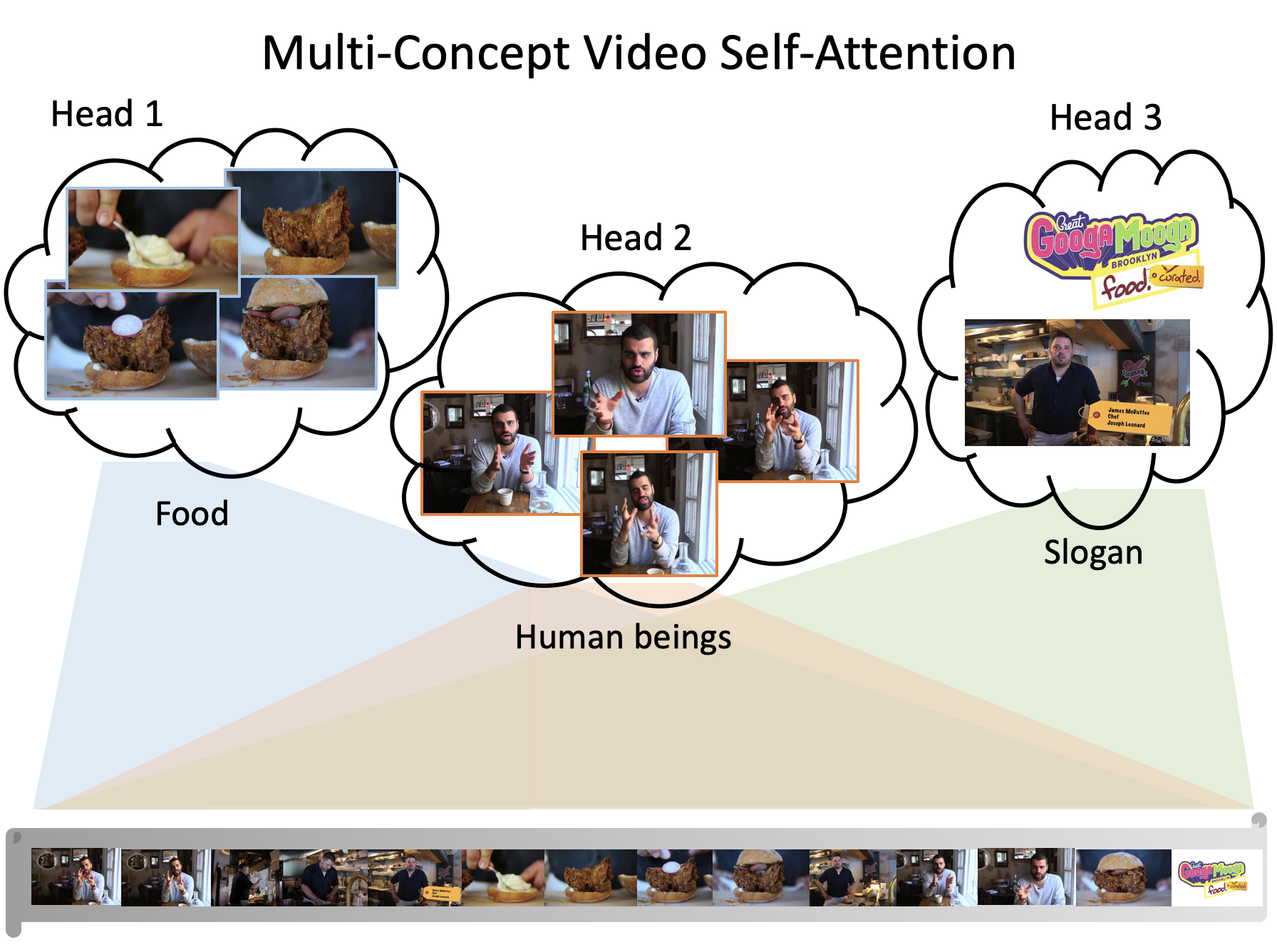}
\end{center}
  \caption{Illustration of advancing self-attention while preserving visual diversity for video summarization. Noted that Head~$1$ to~$3$ denote attention performed in a subspace, which describes proper visual concept information across frames. We proposed a multi-concept video self-attention (MC-VSA) framework for solving this task.}
  \vspace{-8.0mm}
\label{fig:picture1}
\end{figure}

%% file: 2_related_work.tex
\section{Related Work}
\label{sec:Related}
\paragraph{\textbf{Video summarization.}}
Video summarization is among active research topics in computer vision. Several deep methods~\cite{zhang2016video,jung2018discriminative,mahasseni2017unsupervised,zhang2018retrospective,zhou2018deep} developed for video summarization choose to employ long short-term memory (LSTM) cells~\cite{LSTM1997}. For instance, \cite{zhang2016video} consider video summarization as a key-frame/shot selection task, and propose an LSTM-based model for addressing this task. Since most of the videos contain hundreds even thousands of frames, it might not be easy for LSTMs to handle such long-term temporal dependency of videos. Hence, some existing approaches~\cite{zhao2017hierarchical,zhao2018hsa,rochan2018video} are further developed to deal with long videos. \cite{zhao2017hierarchical,zhao2018hsa} propose a hierarchical structure of RNN to exploit intra and inter-shot temporal dependency via two LSTM layers, respectively. Such a hierarchical structure is considered to be more preferable for handling video data with hundreds of frames. On the other hand, \cite{rochan2018video} develop fully convolutional networks for video summarization which requires less training time due to the use of parallel computation techniques. Nevertheless, solving video summarization problems typically requires one to consider the importance across video frames. Existing models generally view the contributions of each frame are equally important during their training stage, which might limit the summarization performance.

\input{Figures/2_model.tex}

\paragraph{\textbf{Attention based summarization.}}
Attention-based models~\cite{ji2017attention,ji2017video,fu2019attentive} have been proposed to video summarization tasks with the goal to alleviate the aforementioned problems. For example, \cite{ji2017attention} introduce an attention mechanism for detecting the shot boundaries, aiming at improving the summarization performances. An attentive encoder-decoder framework is presented in~\cite{ji2017video}, with the models AVS to video summarization via matrix addition and multiplication techniques. \cite{fu2019attentive} utilize adversarial learning for visual attention with models of~\cite{vinyals2015pointer}, which aims at learning a discriminator to detect highlighted fragments as a summery in the input video. Yet, these attention-based methods typically require ground-truth highlighted supervision, and thus it might not be easy to extend to the cases when such labels are not available.

\paragraph{\textbf{Semi-supervised and unsupervised summarization.}}
In order to overcome the above concerns,  unsupervised~\cite{mahasseni2017unsupervised} and semi-supervised~\cite{zhang2018retrospective} method have been proposed. \cite{mahasseni2017unsupervised} is the first unsupervised deep learning paper for video summarization, which uses GAN$_{dpp}$~\cite{mahasseni2017unsupervised} with different LSTM modules to select key-frames from the input video via adversarial learning. On the other hand,~\cite{zhang2018retrospective} use an encoder-decoder mechanism aim at enforcing similarity between the input and the summarized outputs. Nevertheless, the above models take two different LSTM module to maintain the information consistency between raw video and the summary, which cannot ensure the video information is in the embedding from the LSTM module. Our MC-VSA model uses share-weighted LSTM module to encode the video feature and attended feature (resulting the summary later) enforcing the embedding from the LSTM encoder represents the semantic meaning about the video.

%% file: Figures/2_model.tex
\begin{figure*}[t]
    \centering
    \includegraphics[width=\linewidth]{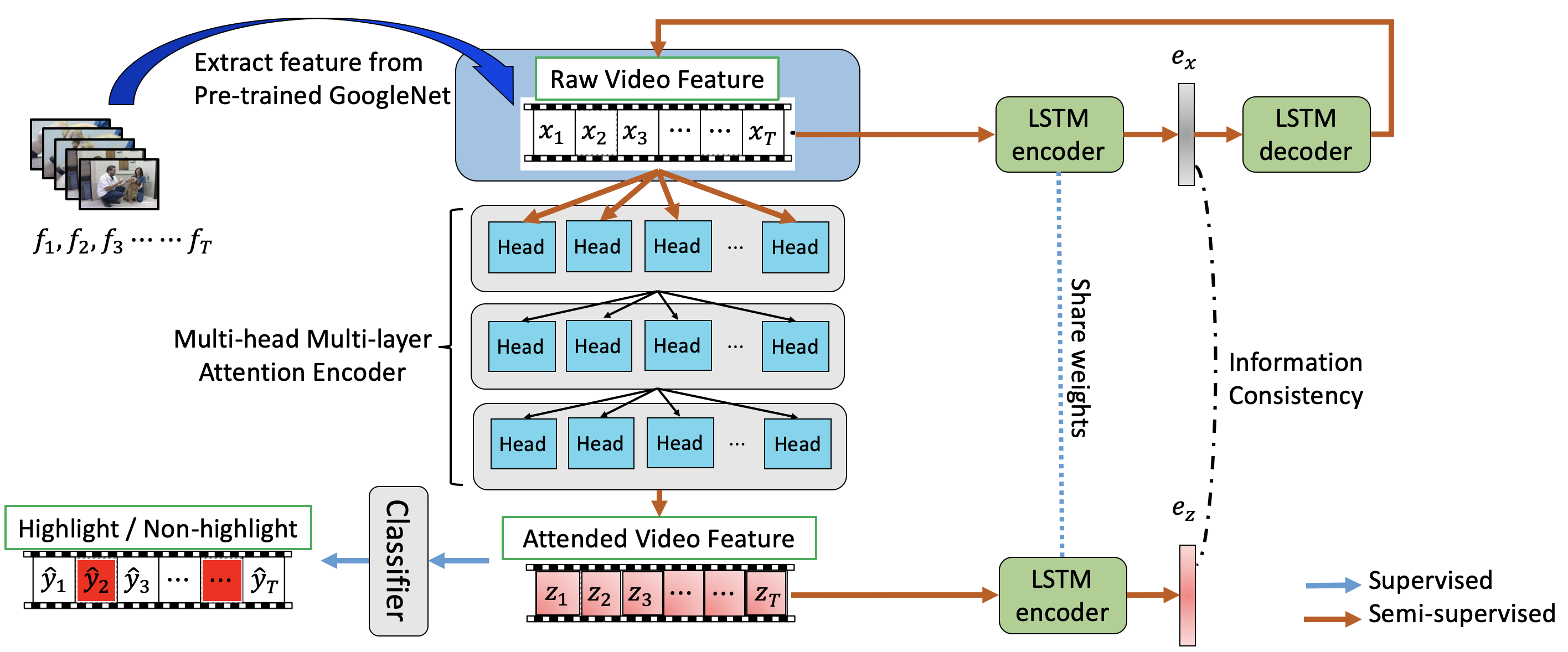}
    \caption{Overview of our Multi-Concept Video Self-Attention (MC-VSA) for video summarization. Our MC-VSA is composed of three modules: the multi-head multi-layer attention encoder, classifier, and the LSTM-based auto-encoder module. Note that the attention encoder takes input videos $X = \{x_i\}_{i=1}^T$ and transforms them into attended features $Z = \{z_i\}_{i=1}^T$, followed by the classifier to output the associated highlight scores $\hat{y}_i$. The LSTM-based auto-encoder module preserves data representation ability while enforcing visual concept similarity, allowing guidance of self-attention for summarization purposes.
    }
    \vspace{-3.0mm}
    \label{fig:picture2}
\end{figure*}

%% file: 3_method.tex
\section{Proposed Method}
\label{sec:Method}

\subsection{Problem Formulation and Notation}
\label{subsec:definition}
Given an input video with a total of $T$ frames, our goal is to select the most important key-frames, about 15\% of the whole video, as the summarization output. We now define the notations to be used in this paper. Assume that we have frame set $F = \{f_i\}_{f=1}^T$ with the associated label set $Y = \{y_i\}_{i=1}^{T}$, where $f_i \in \bbR^{H \times W \times 3}$ and $y_i \in \bbR$ represent the $i^{th}$ frame in the target video. To extract the visual features from the frame set $F$, we apply a CNN  (pre-trained on ImageNet) and obtain the video feature set $X = \{x_i\}_{i=1}^T$, where $x_i \in \bbR^{d}$ ($d$ denotes the dimension of the visual feature).

\subsection{Overview of MC-VSA}
\label{subsec:global_view}
As shown in Figure~\ref{fig:picture2}, we propose a Multi-Concept Video Self-Attention model (MC-VSA) to identify the most representative frames for summarization purposes. Our Multi-Concept Video Self-Attention model is composed of three modules: the multi-head multi-layer attention encoder, classifier, and the LSTM auto encoder decoder module. First, the model takes $X = \{x_i\}_{i=1}^T$ with $T$ sequential features as input of its attention encoder. Our attention encoder then transforms input features $x_i$ in $X$ into different subspaces where attention can be performed accordingly. As stated earlier, the attention encoder allows one to exploit various visual appearances during the attention process, and thus identify concept informative regions across frames. 

We note that, while the learning of MC-VSA can be simply trained using unlabeled data, we further introduce the visual concept loss, which would guide the MC-VSA if the input video data is with ground-truth highlighted labels. To be more precise, we encourage the learned attention weighted features $Z = \{x_i\}_{i=1}^{T}$ to preserve the same video information as the original one ($X = \{x_i\}_{i=1}^T$). To achieve this, the shared LSTM encoder in our framework is designed to match latent vectors $e_z$ and $e_x$ from the same video, thus implying visual concept similarity. If ground-truth highlighted labels are available, the final classfier thus takes the attended features $Z = \{z_i\}_{i=1}^T$ to produce the final highlighted scores $\hat{y}_i$ for each $z_i$. With label supervision available, we encourage the output labels $\hat{Y} = \{\hat{y}_i\}_{i=1}^{T}$ to match the corresponding ground truths $Y = \{y_i\}_{i=1}^{T}$. More details about our MC-VSA model are elaborated in the following. 

As for testing, we take the input video with $T^{\prime}$ frames and produce the final summarization outputs from $\hat{Y} = \{\hat{y}_i\}_{i=1}^{T}$. We note that, our MC-VSA is able to handle a different number of frames in a video as input. The experimental results will be presented in the next section.

\subsection{Video Self-Attention for Summarization}
\label{subsec:attention}

The multi-head multi-layer attention encoder in our MC-VSA is inspired by the Transformer~\cite{vaswani2017attention}. To perform concept-temporal attention from the input video, we project the ``temporal'' video features across video frames onto different subspaces. Each sub-space aims at observing distinct visual concepts as verified later. 
To be more specific, this attention encoder module is developed to transform input video frames into $N$ subspaces by the introduced $N$ self-attention heads, with the goal of observing and distilling potentially representative information across video frames. It then aggregates the attended results from each subspace and produces the final attended features, which can be seen as jointly incorporating the temporal and concept information. In additional, we also perform such multi-head attention across image layers to exhibit robustness in identifying representative visual concepts. 

\subsubsection{Standard Self-Attention.}
\label{subsubsec:Naive} 

For the sake of completeness, we briefly review the self-attention module \cite{xu2015show}. Typical self-attention mechanisms transform the input features into three inputs: query~$Q$, key~$K$, and value~$V$ by matrix multiplication with transforming matrix. The softmax layer will take the result of the multiplication of $Q$ and $K$, and produce the attention weights. Hence, the target attention result is produced from the result of the final matrix multiplication of softmax and the $V$.
 
\subsubsection{Multi-Concept Visual Self-Attention for Video Summarization.}
\label{subsubsec:Multiple_visual}

To observe both temporal and concept information from the input video frames $F = \{f_i\}_{i=1}^T$, we advance the idea of multi-head multi-layer self-attention as described below. As depicted in Fig~\ref{fig:picture3}, we have the attention encoder comprise of $N$ self-attention modules (i.e., the head number equals $N$), and each of them is developed to derive the attended feature each of $N$ subspaces. We firstly transform the input $X$ into $N$ subspace by the $N$ projection layers $P_n$ ($\bbR^{d^n} \leftarrow \bbR^{d}$) where $n$ denotes the projection layer number ($n = 1\sim N$) and $d^n$ denotes the subspace dimension.

To produce the finalized attended results from all of the $N$ subspaces, we introduce the linear projection layer $M^R$ to derive the final attended features $R = \{r_i\}_{i=1}^T$, where $r_i \in \bbR^{d}$ (same dimention as $X_i$), for the original input features $X = \{x_i\}_{i=1}^T$, which can be formulated as:
\begin{equation}
    R = M^R\cdot\mathrm{concat}(O_{1:N}),
\end{equation}
where $\mathrm{concat}$ means we concatenate the outputs~$O_{1:N}$ from all of the $N$ self-attention blocks in the subspace.

To extract rich information from video, we employ $L$ layers in the attention encoder as shown in Figure~\ref{fig:picture2}. Namely, the output of $R^{\prime}$ at the first layer will be passed to the second one to produce the fine-grained output $R^{\prime\prime}$. Hence, the finalized attention features $Z = \{z_i\}_{i=1}^T$ is denoted as $Z = R^{(L)}$.

Later in our experiments, we will present example visual self-attention results produced by different heads e.g., Figure~\ref{fig:exp1} confirming that the attention encoder exhibits sufficient capability in exploiting visual appearance variations across video frames for attention.

\input{Figures/3_multi_attention.tex}

\subsection{Self-Learning of Video Semantic Consistency for Video Summarization}
\label{subsec:post-attention}
The aforementioned attention mechanism can be viewed as a self-summarization process, but lack the ability to ensure that the attended outputs produced by the our attention modules would preserve the information in the input video.

To alleviate this limitation, we apply a Siamese network based on a shared LSTM encoder and a single decoder as illustrated in Figure~\ref{fig:picture2}. This shared LSTM encoder aims at deriving the compressed latent vectors $e_z$ and $e_x$ for the attended feature set $Z$ and the original feature set $X$, while the LSTM decoder is to recover the encoded representation for reconstruction purposes. Thus, we have the reconstruction loss $\mathcal{L}_\mathrm{rec}$ observe the output of this auto-encoder module:

\begin{equation}
\label{equ:rec}
    {\cal L}_\mathrm{rec} = \sum_{i=1}^{T} {\left \| {
        \hat{x}_i - x_i
    }\right \|}^2,
\end{equation}
where $\hat{X} = \{\hat{x}_i\}_{i=1}^T$ denotes the reconstructed feature set and $\hat{x}_i$ indicates the $i$th recovered video frame feature.

More importantly, to preserve visual concept consistency, we require the encoded vectors $e_z$ and $e_x$ to be close if they are from the same video input. As a result, we enforce the visual concept consistency loss $\mathcal{L}_\mathrm{con}$ as follows:
\begin{equation}
\label{equ:consis}
\begin{aligned}
   {\cal L}_\mathrm{con} = {\left \| {
        e_x - e_z
    }\right \|}^2. 
\end{aligned}
\end{equation}

It is worth noting that, our reconstruction loss $\mathcal{L}_\mathrm{rec}$ and consistency loss $\mathcal{L}_\mathrm{con}$ are both computed without observing any ground-truth label. That is, the introduction of this module allows training using unlabeled video. Together with the labeled ones, our proposed framework can be learned in a semi-supervised fashion. As later verified in our experiments, this would result in promising video summarization performances when comparing against state of the arts.

\subsection{Full Objectives}
\label{subsec:object}
As depicted in Figure~\ref{fig:picture2}, our proposed framework can be learned with fully labeled video. That is, the classification layer takes the resulted attended feature set $Z = \{z_i\}_{i=1}^T$ to produce the final highlight potential score $\hat{y}_i$ for each attended feature $z_i$. More precisely,
we encourage the output highlight labels $\hat{Y} = \{\hat{y}_i\}_{i=1}^{T}$ produced by our method can be closed to the ground truth $Y = \{y_i\}_{i=1}^{T}$ and the binary cross-entropy classification loss $\mathcal{L}_\mathrm{cls}$ is formulated as below: 
\begin{equation}
\begin{aligned}
\label{equ:sum}
    {\cal L}_\mathrm{cls} = -\cfrac{1}{T} \sum \limits_{t=1}^T
    ~y_t~\log(\hat{y}_t) + (1-y_t)~\log(1-\hat{y}_t).
\end{aligned}
\end{equation}

\noindent Thus, the total loss $\mathcal{L}$ is summarized as:
\begin{equation}
  \begin{aligned}
  \mathcal{L} = \mathcal{L}_\mathrm{cls} + \mathcal{L}_\mathrm{rec} + \mathcal{L}_\mathrm{con},
  \end{aligned}
  \label{eq:fullobj}
\end{equation}
where $\mathcal{L}_\mathrm{cls}$ is calculated by labeled data, while the $\mathcal{L}_{rec}$ and $\mathcal{L}_\mathrm{con}$ are derived by both the labeled and unlabeled ones. 

We note that, if there is no labeled video data is available during training, $\mathcal{L}_{cls}$ in~\eqref{eq:fullobj} cannot be computed. Following \cite{zhou2018deep,rochan2018video}, we train our MC-VSA in such \textit{unsupervised} setting and introduce a diversity loss $\mathcal{L}_\mathrm{div}$ \eqref{equ:div} to~\eqref{eq:fullobj}. This modification would encourage MC-VSA to select informative yet distinct frames with representative information in unsupervised learning scenario.
\begin{equation}
  \begin{aligned}
    {\cal L}_\mathrm{div} = 
    \sum_{s=1}^{S}\sum_{x^s\in{{c}^s}} \mathop{\sum_{{x^s}'\in{{c}^s}}}_{x^s\neq {x^s}'} 
    d(x^s, {x^s}').
  \end{aligned}
  \label{equ:div}
\end{equation}

%% file: Figures/3_multi_attention.tex
\begin{figure}[t]
\begin{center}
\includegraphics[width=0.6\linewidth]{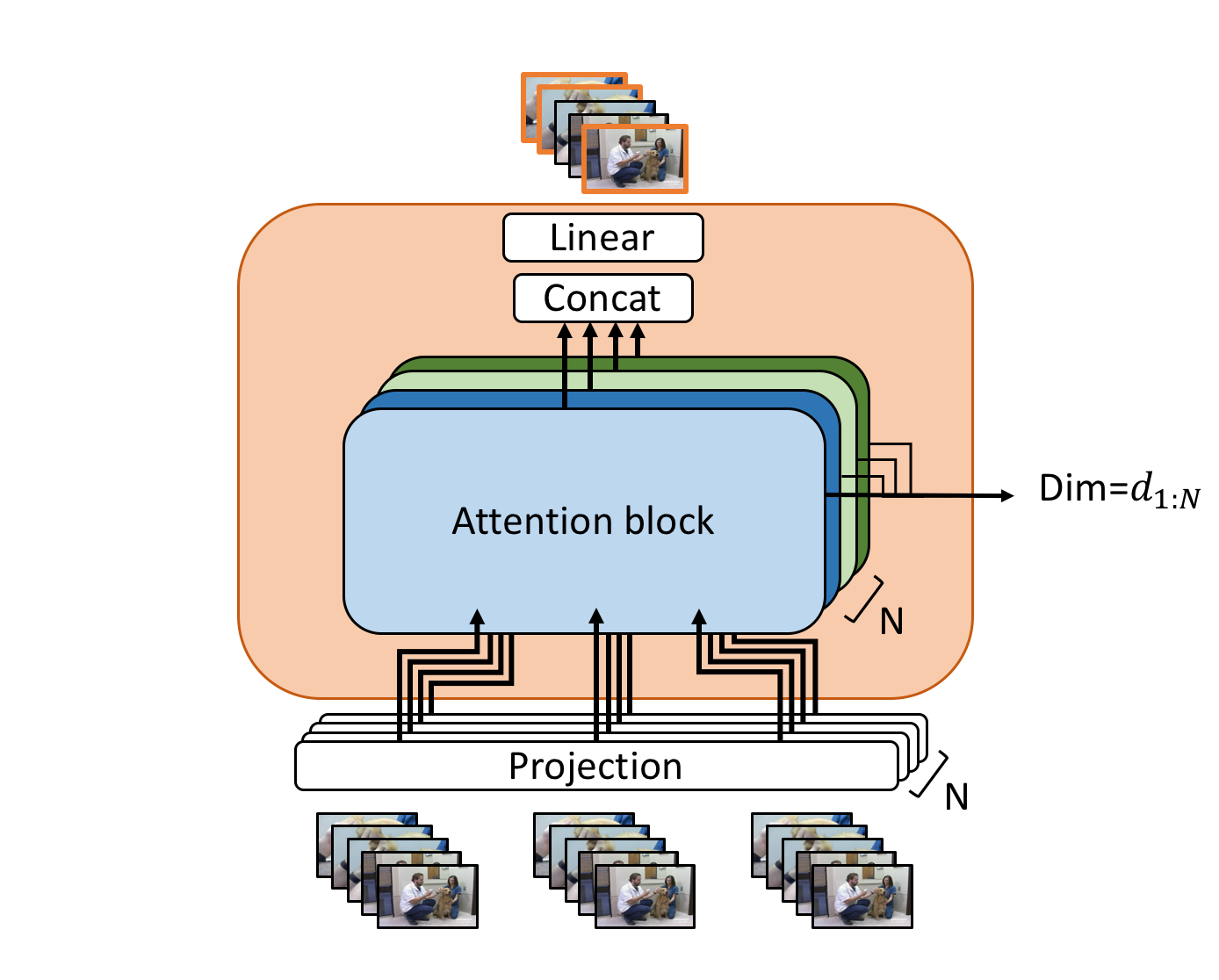}
\vspace{-7mm}
\end{center}
  \caption{Illustration of multi-head multi-layer self-attention module of our attention encoder (note that only a single layer is shown for simplicity). With $N$ different single-head attention blocks (each with a projection matrix layer), self-attention can be performed in different subspaces (dimension $d_n$ for each) for capturing diverse visual concepts. We concatenate the outputs $O_{1:N}$ from all attention blocks and obtain the joint attention result $R$ at the output of the final linear transform layer.}
\label{fig:picture3}
\vspace{-5mm}
\end{figure}

%% file: 4_experiment.tex
\input{Tables/Supervised.tex}
\input{Tables/Unsupervised.tex}
\section{Experiment}
\label{sec:Exp}
In this section, we first describe the datasets in Sec.~\ref{sec:Exp:dataset}, followed by the experiment protocols and implementation details in Sec.~\ref{sec:Exp:imple}. For evaluating our MC-VSA, we present quantitative results in Sec.~\ref{sec:Exp:quan} and Sec.~\ref{sec:Exp:unsuper}. We also provide ablation studies in Sec.~\ref{sec:Exp:ablation}. Finally, we provide qualitative results and visual analysis in Sec.~\ref{sec:Exp:qual}.
\subsection{Datasets}
\label{sec:Exp:dataset}
We evaluate our method on two public benchmark datasets SumMe~\cite{gygli2014creating} and TvSum~\cite{song2015tvsum}, and use additional dataset: OVP and YouTube~\cite{zhou2018deep} in the Augmented and Transfer settings:

\paragraph{\textbf{SumMe.}} SumMe consists of 25 videos with several different topics such as holidays and sports. Each video ranges from 1 to 6 minutes and annotated by 15 to 18 persons. Thus, there are multiple ground truth summaries for each video. 

\paragraph{\textbf{TvSum.}} TvSum is a larger dataset with 50 videos and include topics like news and documentaries. The duration of each video is from 2 to 10 minutes. Same as SumMe, TvSum dataset has 20 annotators providing frame-level importance scores. Following~\cite{zhang2016video} and~\cite{zhou2018deep}, we convert important scores to shot-based summaries for evaluation. 

\paragraph{\textbf{OVP and YouTube.}} Followed by ~\cite{zhang2016video} and~\cite{zhou2018deep}, we consider two additional challenging datasets released by~\cite{zhou2018deep}: OVP and YouTube, which contain 50 videos and 39 videos in the augmented and transfer settings

\input{Figures/4_visual_1.tex}
\subsection{Protocols and Implementation Details}
\label{sec:Exp:imple}
\paragraph{\textbf{Evaluation protocols}}
We follow the three settings adopted in~\cite{zhang2016video,zhou2018deep,rochan2018video} to evaluate our methods:

\begin{itemize}
    \item Canonical: we use the standard supervised learning on the dataset, i.e., $80 \%$ for training and the rest for testing following previous work \cite{zhang2016video,mahasseni2017unsupervised,ji2017video,zhou2018deep,rochan2018video}. 
    \item Augmented: we use the standard supervised training as the canonical setting but augment the training data with OVP and YouTube datasets.
    \item Transfer: We use three datasets as the training data and a target dataset (e.g. SumMe or TvSum) as the testing data to evaluate the transfer ability of our model.
\end{itemize}
\vspace{-2.0mm}
For fair comparison, we follow the commonly adopted metric in previous works~\cite{zhang2016video,zhou2018deep,rochan2018video}, and computed F-score to assess the similarity between automatic and ground-truth summaries. As for the training/testing data, we apply the same standard supervised learning setting as \cite{zhang2016video,mahasseni2017unsupervised,ji2017video,zhou2018deep,rochan2018video} where the training and testing are from the disjoint part of the same dataset. We randomly leave $20\%$ for testing and the remaining $80\%$ for training. We report the results at F-score in all of the settings.
\vspace{-2.0mm}
\paragraph{\textbf{Implementation details}}
We implement our framework using PyTorch. We downsample the video data into frame sequences in $2$ fps as previous work~\cite{zhang2016video,zhou2018deep}. For fair comparisons with \cite{zhang2016video,rochan2018video,zhou2018deep,ji2017video}, we also employ GoogleNet \cite{szegedy2015going} pre-trained on ImageNet as backbone as our CNN for extracting the video features while the output dimension $d$ is $1024$ (output of pool5 layer of the GoogleNet). All of the attention modules is composed of linear projection matrices as mentioned in the Section~\ref{sec:Method}. We set the number of heads $N$ as $24$ while the dimension $d_n$ of each subspace features are set as $\{ 64~| n = 1 \sim 12 \} $ and $\{ 128~| n = 13 \sim 24 \}$. Our MC-VSA comprises of $3$ multi-head attention layer, i.e., we set $L$ as $3$. The classifier is composed of a fully connected layer followed a sigmoid activation. The LSTM encoder and decoder in our model contain $512$ units. Besides, we set the learning rate as $1e^{-4}$ for all of our components. We use Adam optimizer to train the MC-VSA by optimizing the objective loss. We produce the summary outputs by KNAPSACK algorithm following \cite{zhang2016video,rochan2018video}

\input{Tables/Ablation_TvSum.tex}
\subsection{Comparison with Supervised Approaches}
\label{sec:Exp:quan}
We compare our proposed MC-VSA with state-of-the-art methods on two benchmark datasets, and summarize the results in Table~\ref{tab1}.

\paragraph{\textbf{SumMe.}}
For the canonical setting, we see that our MC-VSA performed favorably against recent LSTM based approaches (e.g., Bi-LSTM~\cite{zhang2016video}, DPP-LSTM~\cite{zhang2016video}, GAN$_sup$~\cite{mahasseni2017unsupervised}, and DR-DSN$_sup$~\cite{zhou2018deep}) by a large margin e.g., $9.5\%$, $9\%$, $7.4\%$, and $5.6\%$. For the CNN-based model of SUM-FCN~\cite{rochan2018video}, we observe our performance gain of $4.1\%$ in terms of F-scores. We also compare our model with the hierarchical LSTM model H-RNN \cite{zhao2017hierarchical} and HSA-RNN \cite{zhao2018hsa} at f-score by $7.3\%$ and $7.5\%$ and attention model M-AVS~\cite{ji2017video} and VASNet~\cite{ji2017attention} by $7.2\%$ and $1.9\%$. For both augment and transfer settings, we also observe similar trends and achieve improved performances against state-of-the-art methods. 

\paragraph{\textbf{TvSum.}}
In canonical setting, our model achieves the improvement over the standard LSTM module Bi-LSTM~\cite{zhang2016video}, DPP-LSTM~\cite{zhang2016video}, GAN$_sup$~\cite{mahasseni2017unsupervised}, and DR-DSN$_sup$~\cite{zhou2018deep} ) by a large margin $14\%$, $13\%$, $9.9\%$, and $9.5\%$. Compared with SUM-FCN, $8.9\%$ improvement at f-score is observed. For previous hierarchical LSTM module and attention model, we all have improved summarization score. Similarly in augment and transfer setting, our model performs favorably against the state-of-the-arts.
\input{Figures/6_visual_3.tex}

\vspace{-4.0mm}
\subsection{Comparisons with Unsupervised Approaches}
\label{sec:Exp:unsuper}
We report our unsupervised learning results and comparisons in Table~\ref{tab2}. With training strategies presented in Sect.~\ref{sec:Method}, we evaluate the effectiveness of our MC-VAS in the unsupervised setting by comparing with five existing unsupervised summarization methods~\cite{mahasseni2017unsupervised,zhou2018deep,rochan2018video,jung2018discriminative}. As shown in Table~\ref{tab2}, our MC-VAS was able to achieve comparable results with the state-of-the-arts did on both SumMe and TvSum. Thus, even without any supervision, we can confirm that our model takes advantages from multi-concept video self-attention with visual concept consistency for video recovery and summarization.

\input{Figures/5_visual_2}

\vspace{-4.0mm}
\subsection{Qualitative Results}
\label{sec:Exp:qual}
To analyze the effectiveness of the self-attention module in MC-VSA, we present visualization results in Fig.~\ref{fig:exp1}, in which the attention outputs were observed from the second (high-level) layer in our model. 
In Fig.~\ref{fig:exp1}, the upper half part illustrates frame-level attention weights for the selected $13$ heads in our model. Note that each row in the upper part of this figure represents a single head, in which the darkness of the color indicates the importance of the associated frame. From this example result, we see that the attention weights for different heads are quite different, which confirms that our multi-head self-attention mechanism leads to visual concept diversity.  
For example, by comparing the learned attention weights and the corresponding frames (i.e., upper vs. lower parts of Fig.~\ref{fig:exp1}), we see that one head in blue rectangular box exhibits the semantic meaning of hamburger, while the red one indicates the appearance of the food critic. And, as confirmed by earlier quantitative experiments, these resulting attention weights across different heads are indeed correlated with the summarization outputs.

On the other hand, Fig.~\ref{fig:exp3} illustrates the attention observed in the first layer of our MC-VSA, which can be viewed as low-level self-attention of multiple heads from the input video. Take the entry of the $i$th column at the $j$th row, its value reflects the attention for the corresponding frame pair. From this figure, we see that the attention boundaries were visible and generally matched the shot boundaries. IN addition, we see that visual concepts with similar visual appearances (e.g., wheel, car, etc.) were properly identified in particular video segments, which reflect the concept-specific video shot information of this input video. Guided by the classification and data recovery losses, this explains why our proposed model is able to capture multiple representative visual information, achieving satisfactory summarization outputs.
\vspace{-4.0mm}
\subsection{Ablation Studies}
\label{sec:Exp:ablation}
\paragraph{\textbf{Semi-supervised settings.}}
We first conduct semi-supervised learning analysis of our proposed MC-VSA on the TvSum dataset. As illustrated in Figure~\ref{fig:exp2}, the vertical axis indicates the F-score, and the horizontal axis represents the percentage of the labeled data observed during training. 
For the completeness of analysis, we compare our approach with $5$ supervised or unsupervised summarization methods in the same figure: Bi-LSTM \cite{zhang2016video}, DPP-LSTM \cite{zhang2016video}, GAN$_sup$~\cite{mahasseni2017unsupervised}, DR-DSN$_sup$~\cite{zhou2018deep}, and SUM-FCN \cite{rochan2018video}, and $2$ unsupervised methods: GAN$_dpp$~
\cite{mahasseni2017unsupervised}, DR-DSN \cite{zhou2018deep}. From the results presented in the figure, we see that our MC-VSA achieved improved performances over others. Moreover, we see that our method was able to perform favorably against existing supervised approaches by a large margin, even when only $25\%$ labels were observed by our model during training. 
Furthermore, Figure~\ref{fig:exp2} compares our model with its variants in semi-supervised settings. Note that Ours* denotes our model excluding both reconstruction loss and visual concept consistency loss. Refer to the semi-supervised analysis, the performance drop between Ours and Ours* confirms that such two loss terms are crucial when observing unlabeled data. We note that, in figure~\ref{fig:exp2} for cases 100\%, the performances achieved by ours and ours* respectively are the same. 
This is because when we use entire label set in 100 \%, LSTM module only serve to train our model more stable instead of achieving improved performance.

\input{Figures/7_visual_4}
\paragraph{\textbf{Network architecture design.}}
We now further discuss the design of our model architecture.
In Figure~\ref{fig:exp4}, we show the performance of our multi-headmulti-layer  attention  model  with  varying  numbers  of  layers and heads (x-axis).  From this figure, we see that while such hyperparameters need to be determine in advance, the results were not sensitive to their choices.  In other words, with a sufficient number of heads and layers, multiple visual concepts can be extracted for summarization purposes as shown in our supplementary video. As shown in Table~\ref{tab}, we apply three evaluation protocols, including F1, Kendall's~$\tau$, and Spearman's~$\rho$, to evaluate our MC-VSA model. Kendall's~$\tau$, and Spearman's~$\rho$ are proposed by \cite{re2019} for impartial comparison. We compare our full model (3layers-24heads) with other baseline models. To be more specific, we take the VASNet~\cite{ji2017attention} as the naive self-attention model baseline. Ours (w/o attention) represents the MC-VSA model consisting of only classifier while ours (1layer-1head) indicates only single layer and head in the attention encoder. The performance drop is observed when comparing ours with the above mentioned baseline models. We additionally report the performance provided by \cite{zhang2016video} and \cite{zhou2018deep} in \cite{re2019} in Kendall's~$\tau$ and Spearman's~$\rho$ evaluation protocol for benchmark comparison.

%% file: Tables/Supervised.tex
\begin{table}[t] 
    \centering
    \caption{
    Comparisons with existing supervised summarization methods on SumMe and TvSum in differnt experimental settings. The numbers in bold and under line indicate the best and the second result. 
    }
    \resizebox{0.6\linewidth}{!}{
    \begin{tabular}{l|ccc|ccc} 
    \toprule
    \multirow{2}{*}{Method} & \multicolumn{3}{c|}{SumMe} & \multicolumn{3}{c}{TvSum}\\ 
     &C&A&T &C&A&T  \\
    \midrule
    Bi-LSTM~\cite{zhang2016video} 
    &37.6 &41.6 &40.7   &54.2 &57.9 &56.9 \\
    DPP-LSTM~\cite{zhang2016video}
    &38.6 &42.9 &41.8   &54.7 &59.6 &58.7 \\
    GAN$_{sup}$~\cite{mahasseni2017unsupervised} 
    &41.7 &43.6 &-      &56.3 &61.2 &-    \\
    DR-DSN$_{sup}$~\cite{zhou2018deep}
    &42.1 &43.9 &42.6 &{58.1} &59.8 &{58.9}\\
    SUM-FCN~\cite{rochan2018video}  
    &47.5 &51.1 &{44.1}   &56.8 &59.2 &58.2 \\
    re-SEQ2SEQ~\cite{zhang2018retrospective}
    &44.9 &- &- &{\bf63.9} &- &-\\
    UnpairedVSN~\cite{rochan2019video}
    &47.5 &- &41.6 &55.6 &- &55.7\\
    \midrule
    H-RNN~\cite{zhao2017hierarchical}
    &44.3 &- &- &62.1 &- &-\\
    HSA-RNN~\cite{zhao2018hsa}
    &44.1 &- &-  &59.8 &- &-\\
    \midrule
    M-AVS~\cite{ji2017video}
    &44.4 &46.1 &- &61.0 &61.8 &-\\
    VASNet~\cite{ji2017attention}
    &49.7 &51.1 &- &61.4 &62.4 &-\\
    \midrule
    MC-VSA (Ours)
    &{\bf51.6} &{\bf53.0} &{\bf48.1}  &{\underline{63.7}} &{\bf64.0} &{\bf59.5}\\
    \bottomrule
    \end{tabular}
    }
    \vspace{-5.0mm}
    \label{tab1}
\end{table}

%% file: Tables/Unsupervised.tex
\begin{table*}[t] 
    \centering
    \caption{
    Comparisons with recent unsupervised approaches for video summarization using SumMe and TvSum. Note that * indicates the non deep-learning based methods. The number in bold indicates the best performance.
    }
    \resizebox{0.6\linewidth}{!}{
    \begin{tabular}{l|c|c| c|c|c|c} 
    \toprule
    {DATASET} 
    &\cite{song2015tvsum}*
    &\cite{mahasseni2017unsupervised}&\cite{rochan2018video}&\cite{zhou2018deep}&\cite{rochan2019video}& MC-VSA (Ours) \\
    \midrule
    SumMe& 
    26.6& 39.1& 41.5& 41.4 &{\bf47.5}& {\underline{44.6}}
    \\ 
    \midrule
    TvSum& 
    50.0& 51.7& 52.7& 57.6 &55.6 & {\bf58.0}
    \\
    \bottomrule
    \end{tabular}
    }
    \label{tab2}
    \vspace{-3.0mm}
\end{table*}

%% file: Figures/4_visual_1.tex
\begin{figure*}[t]
    \vspace{-3.0mm}
    \includegraphics[width=1\linewidth]{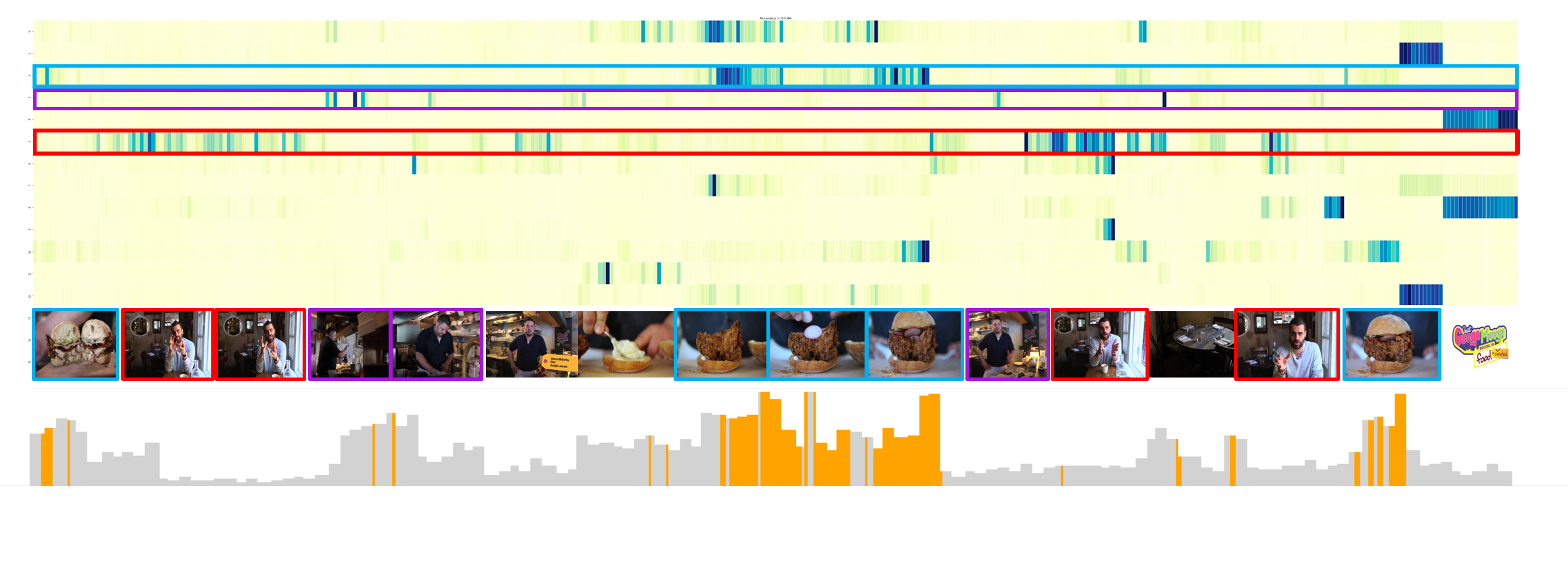}
    \vspace{-5.0mm}
    \caption{Visualization example of our MC-VSA for video summarization on TvSum. We visualize selected attention maps generated by the second layer in the attention encoder, with ground truth (grey) and predicted outputs (orange) summarization shown below. Note that the attention outputs bounded in different colors (blue, red and purple) correspond to different multi visual concepts (e.g., burger, commentator, and chef) in this video.
    }
    \vspace{-5.0mm}
    \label{fig:exp1}
\end{figure*}

%% file: Tables/Ablation_TvSum.tex
\begin{table}[t] 
    \centering
    \caption{
    Ablation studies and performance comparisons on TvSum dataset. We take dppLSTM \cite{zhang2016video} and \cite{zhou2018deep} for comparisons in  Kendall's~$\tau$, and Spearman's~$\rho$ evaluation protocol. The number in bold denotes the best performance.
    }
    \vspace{+5pt}
    \label{table}
    \resizebox{0.7\linewidth}{!}{
    \begin{tabular}{l|c|cc} 
    \toprule
    \multirow{2}{*}{Method} & {w/ knapsack algo.} &\multicolumn{2}{c}{w/o knapsack algo.}\\
    &{F1 score} & {Kendall's~$\tau$}& {Spearman's~$\rho$}\\ 
    \midrule
    dppLSTM
    &60.0 &0.042 &0.055 \\
    DR-DSN$_{dpp}$
    &58.0 &0.020 &0.026 \\
    \midrule
    VASNet
    &61.4 &- &- \\
    Ours (w/o attention)
    &59.7 &0.005 &0.006 \\
    Ours (1layer-1head)
    &60.1 &0.065 &0.079 \\
    \midrule
    Ours (3layers-24heads)
    &{\bf63.7} &{\bf0.116} &{\bf0.142}\\
    \bottomrule
    \end{tabular}
    }
    \vspace{-5.0mm}
    \label{tab}
\end{table}

%% file: Figures/6_visual_3.tex
\begin{figure*}[t]
    \centering
    \includegraphics[width=0.9\linewidth]{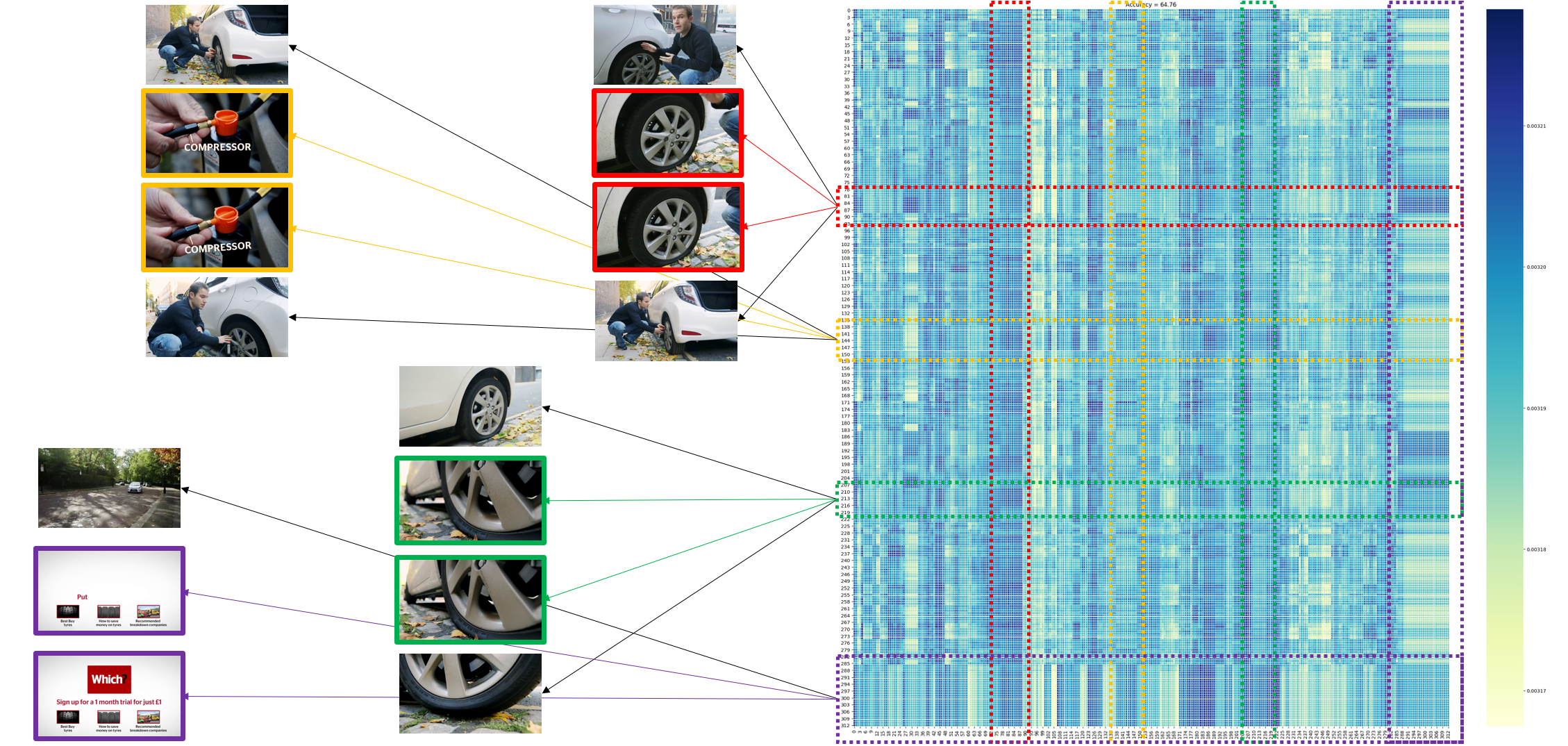}
    \caption{
    Visualization for multi-head self-attention at the first layer of MC-VSA. We show that this low-level attention layer not only implies shot-level attention, visual concepts associated with similar objects are properly attended across video frames (e.g., attention outputs bounded in difference colors).
    }
    \label{fig:exp3}
    \vspace{-5.0mm}
\end{figure*}

%% file: Figures/5_visual_2.tex
\begin{figure}[t]
    \centering
    \includegraphics[width=.8\linewidth]{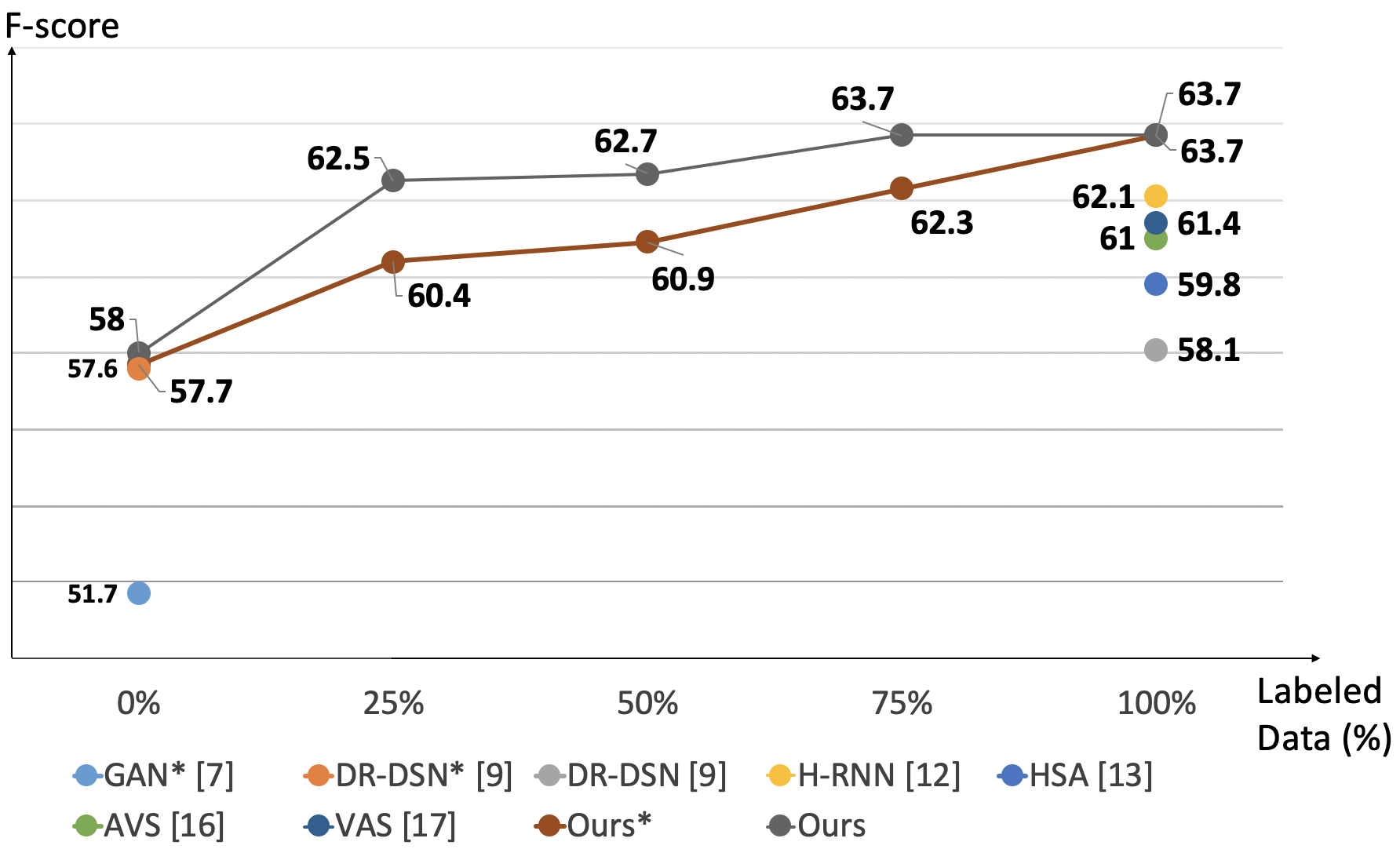}
    \vspace{-3.0mm}
    \caption{Performance analysis of our model in semi-supervised settings on TvSum. The x/y-axis indicate the percentage of labels in the training set and the F-score, respectively. Results of recent supervised and unsupervised approaches are depicted for comparison purposes. Note that Ours* denotes our MC-VSA excluding the LSTM auto-encoder module while Ours represents the full model.
    }
    \vspace{-6.0mm}
    \label{fig:exp2}
\end{figure}

%% file: Figures/7_visual_4.tex
\begin{wrapfigure}{r}{0.6\textwidth}
\vspace{2.0mm}
    \begin{center}
        \includegraphics[width=\linewidth]{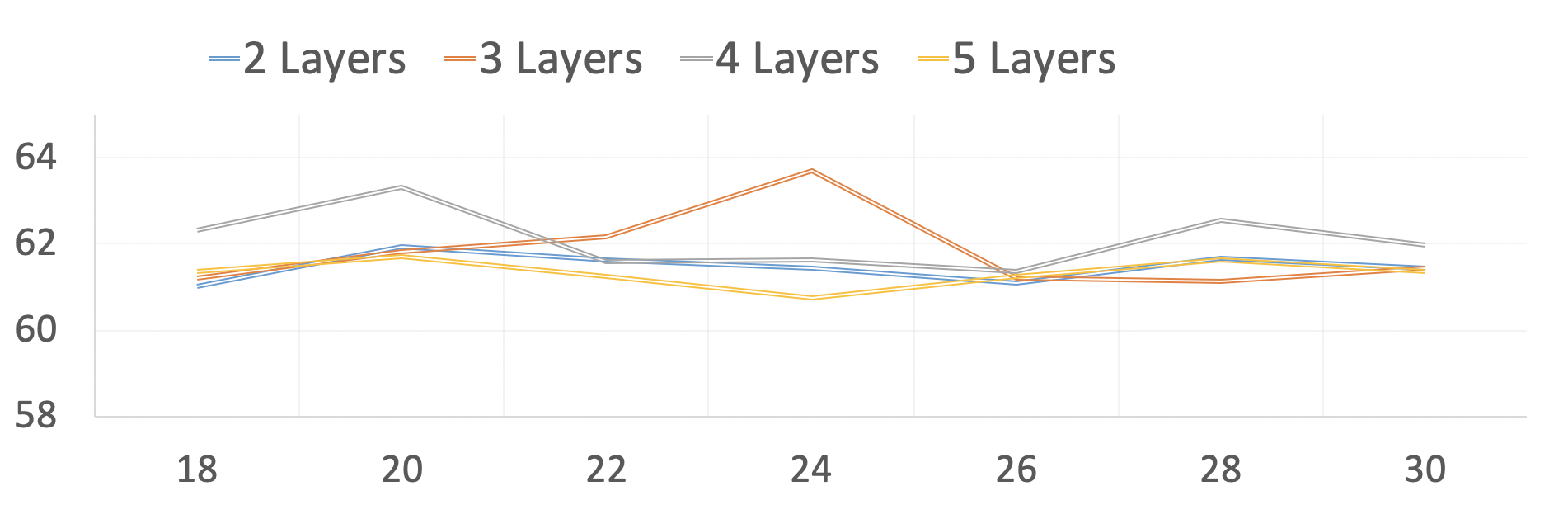}
    \end{center}
\vspace{-5.0mm}
\caption{Performance of our attention model with varyingnumbers of layers (L) and heads (N) (i.e., x-axis). We fix L= 3 and N = 24 for our model in all experiments.}
\label{fig:exp4}
\end{wrapfigure}

%% file: 5_conclusion.tex
\vspace{-4.0mm}
\section{Conclusion}
\label{sec:conclusion}
We presented a novel deep learning framework \emph{multi-concept video self-attention (MC-VSA)} and consistency constraint between the input video and the output summary for video summarization. The core technical novelty lies in the unique design of multi-concept visual self-attention model, which jointly exploits concept and temporal attention diversity in the input videos, while enforcing the summarized outputs to have consistency with original video. Our proposed framework not only generalized in supervised, semi-supervised and unsupervised settings but also in both evaluation protocols. Also, our experiments and qualitative results confirmed the effectiveness of our proposed model and its ability to identify certain informative visual concepts. 
\newpage
